\documentclass[conference]{IEEEtran}
\IEEEoverridecommandlockouts
\usepackage[numbers]{natbib}
\usepackage{amsmath,amssymb,amsfonts}
\usepackage{algorithmic}
\usepackage{graphicx}
\usepackage{textcomp}
\usepackage{hyperref}
\usepackage{balance}
\DeclareRobustCommand{\textsupsub}[2]{{%
  \m@th\ensuremath{%
    ^{\mbox{\fontsize\sf@size\z@#1}}%
    _{\mbox{\fontsize\sf@size\z@#2}}%
  }%
}}

\usepackage{xcolor}
\def\BibTeX{{\rm B\kern-.05em{\sc i\kern-.025em b}\kern-.08em
    T\kern-.1667em\lower.7ex\hbox{E}\kern-.125emX}}

\title{Action similarity judgment based on kinematic primitives\\
\thanks{This work was financially supported by the Knowledge Foundation, Stockholm, under SIDUS grant agreement no. 20140220 (AIR, Action and intention recognition in human interaction with autonomous systems). A.S. is supported by a Starting Grant from the European Research Council (ERC) under the European Union’s Horizon 2020 research and innovation programme. G.A. No 804388, wHiSPER.}
}
\makeatletter
\newcommand*\titleheader[1]{\gdef\@titleheader{#1}}
\AtBeginDocument{%
  \let\st@red@title\@title
  \def\@title{%
    \bgroup\normalfont\large\centering\@titleheader\par\egroup
    \vskip1.5em\st@red@title}
}
\makeatother
\titleheader{\textit{\small{(Preprint) -- 10th Joint IEEE International Conference on Development and Learning and Epigenetic Robotics (ICDL-EpiRob 2020)}}}
\author{
    \IEEEauthorblockN{Vipul Nair\IEEEauthorrefmark{1}\textsuperscript{1}, Paul Hemeren\textsuperscript{1},
    Alessia Vignolo\textsuperscript{2}, Nicoletta Noceti\textsuperscript{3}, Elena Nicora\textsuperscript{3},\\
    Alessandra Sciutti\textsuperscript{2}, Francesco Rea\textsuperscript{4}, Erik Billing\textsuperscript{1}, Francesca Odone\textsuperscript{3}, Giulio Sandini\textsuperscript{4}}\\
    
    \IEEEauthorblockA{\textsuperscript{1}School of Informatics, University of Sk\"ovde, Sweden\\}
    \IEEEauthorblockA{\textsuperscript{2}CONTACT Unit, Istituto Italiano di Tecnologia, Italy\\}
    \IEEEauthorblockA{\textsuperscript{3}MaLGa Center - DIBRIS, Universit\`a di Genova, Italy\\}
    \IEEEauthorblockA{\textsuperscript{4}RBCS Unit, Istituto Italiano di Tecnologia, Italy\\
    \IEEEauthorrefmark{1}vipul.nair@his.se}
}

\begin{document}
\maketitle

\begin{abstract}
Understanding which features humans rely on --  in visually recognizing action similarity is a crucial step towards a clearer picture of human action perception from a learning and developmental perspective. In the present work, we investigate to which extent a computational model based on kinematics can determine action similarity and how its performance relates to human similarity judgments of the same actions. To this aim, twelve participants perform an action similarity task, and their performances are compared to that of a computational model solving the same task. The chosen model has its roots in developmental robotics and performs action classification based on learned kinematic primitives. The comparative experiment results show that both the model and human participants can reliably identify whether two actions are the same or not. However, the model produces more false hits and has a greater selection bias than human participants. A possible reason for this is the particular sensitivity of the model towards kinematic primitives of the presented actions. In a second experiment, human participants' performance on an action identification task indicated that they relied solely on kinematic information rather than on action semantics. The results show that both the model and human performance are highly accurate in an action similarity task based on kinematic-level features, which can provide an essential basis for classifying human actions.

\end{abstract}


\section{Introduction}

Human vision is highly sensitive to the biological motion patterns created by the movement of other individuals (e.g., \cite{tversky2019mind, yovel2016recognizing}). From a developmental perspective, this sensitivity in terms of visual preference is present in newborns \cite{simion2008predisposition} and significantly increases longitudinally from 3 to 24 months \cite{sifre2018longitudinal}.  Learning to distinguish between different action categories and exemplars reflects this sensitivity and visual preference \cite{hemeren2008mind}.

 Judging action similarity, i.e., judging whether two actions are the same or not, is an essential part of learning action categories and a step towards action understanding. Indeed, in most behavioral studies, it has been addressed as a form of measure toward understanding action semantics \cite{watson2014uncovering}, action prototypes \cite{giese2002measurement}, and imitation \cite{catmur2013you}.
 
From a computational viewpoint, judging action similarity is paramount in social robotics, industrial-robot collaboration, and video surveillance \cite{kong2018human}. Action similarity can be complicated in a realistic setting, such as action class ambiguity in multi-class action recognition \cite{qin2015compressive}. To address this ambiguity problem, action similarity labeling (same or different) was first introduced by \citeauthor{kliper2011action} \cite{kliper2011action} as a critical task in action recognition. According to \citeauthor{kliper2011action} \cite{kliper2011action}, action similarity labeling aims to determine if the actors in two video sequences are performing the same or different actions. The labeling algorithms rely primarily on creating a suitable metric for the differences between the actions from the extracted kinematic features (see \cite{qin2015compressive} for a detailed review of the approaches). \citeauthor{kliper2011action} showed a considerable gap (around 65\%) between the state-of-the-art methods and the success rate of humans on action similarity labeling and argued towards a principled understanding of what makes actions similar or different \cite{kliper2011action}. 

The work presented in this paper attempts to reduce this gap by using a computational model that derives action primitives based on kinematic features (from the biological motion regularities) \cite{vignolo2016complexity}. The model is used to perform an action similarity task (AST), i.e., to judge whether actions are the same or different. The model performs AST by learning to classify actions (using dictionary learning) based on a linear combination of kinematic primitives (sparse coding technique). In particular, we assess how (the extent) this representation of actions can produce successful action classification -- by comparing the model's performance with human visual performance based on the same AST.

As a further comparison between the model and human biological motion perception, we conducted a second experiment with an action identification task (AIT) to validate the use of kinematic features in action similarity judgments made by humans. In other words, are humans relying on high-level semantic features for their similarity judgments rather than on low-level kinematics?

Previous studies have shown that humans identify action primitives based on kinematic features in an action segmentation task \cite{hemeren2011deriving}. Consistent with these results, the model derives and uses combinations of different visual body motion patterns (action primitives) to distinguish between different human actions. Besides, action primitives is an area of focus in modeling the recognition and categorization of human actions by artificial systems \cite{kulic2011learning}.

In summary, this paper addresses three questions. What is the extent to which the computational model based on kinematic primitives can determine action similarity among a group of actions? To what extent does the model's performance relate to human similarity judgments of the same actions? Do the human action similarity judgments rely mainly on the kinematic features of the actions rather than higher-level action semantics?


\section{Hand action stimuli}

The stimuli used in this study are taken from Multimodal Cooking Actions (MoCA) \footnote{The dataset is available for download at \url{https://github.com/nicolettanoceti/CookingDataset.}} dataset. The full dataset includes motion capture data, and videos (from multiple viewpoints) of upper body actions executed by one actor in a cooking scenario. (For more details about the dataset, see \cite{malafronte2017investigating}.) The actions are hand based and manipulative, i.e., actions intended to modify or displace an object. This dataset was chosen for testing action similarity as hand-based actions cover a wide range of complexity with various movements, and most day-to-day activities involve hand actions.

\begin{figure}[ht]
\begin{center}
\includegraphics[width=\linewidth]{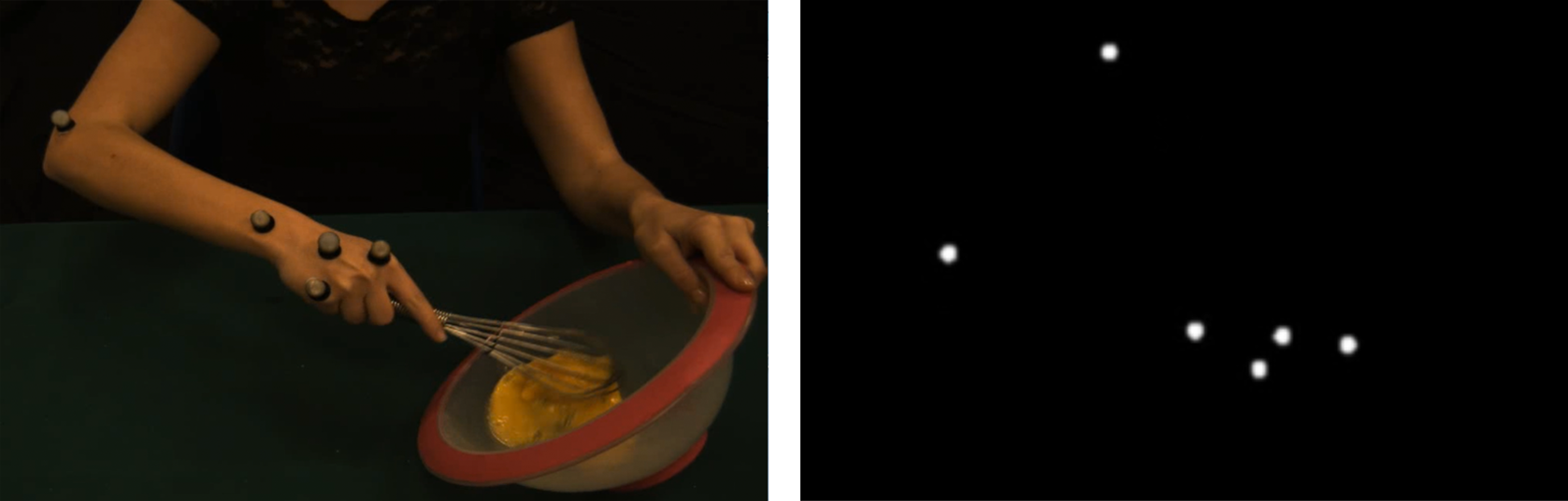}
\caption{The left image shows a frame of action \textit{Eggs} from a frontal point of view, and the right image shows its PLD. The PLDs correspond to the positions of the markers.}
\label{fig:stimuli}
\end{center}
\end{figure}

For this study we chose 19 actions from the set; namely \textit{Carrot: Grating a carrot, Cut: Cutting a loaf of bread, Dish: Cleaning a dish, Eat: Eating a slice of bread, Eggs: Beating eggs, Lemon: Squeezing a lemon, Mezzaluna: Using a mezzaluna knife, Mixing: Stirring a mixture, OpenBottle: Opening a bottle, Pan: Pan flip, Pestare: Crushing leaves, Pouring: Pouring water, Reaching: Reaching an object, Rolling: Rolling dough, Salad: Rotating salad chopper, Salt: Using a salt shaker, Spread: Spreading cheese on bread, Table: Cleaning table, and Transport: Transporting an object} (all the actions will be referred to by their capitalized term).  Most of the actions are carried out by the right hand, whereas some involve both hands (e.g., \textit{Mezzaluna} or \textit{Rolling}).  To investigate action similarity, selecting a single viewpoint was necessary to avoid the excessive duration of the experiment with human participants. Therefore we opted for the frontal viewpoint, which is familiar and natural for interaction, especially during the early stages of child development. However, it has been shown that the model can perform action recognition with multiple viewpoints \cite{VignoloICPR2020}, paving the way for future investigation of human perception. See Fig. \ref{fig:stimuli} to see an example frame for \textit{Eggs}, and its point-light display (PLD). Since this study focuses just on the low-level kinematic features of actions, the human participants were shown PLDs limiting their leverage from contextual information. Alternatively, the model was designed to extract only kinematic information directly from the videos (see Section  \ref{computationalmodel} for details).

\section{Computational model}
\label{computationalmodel}
The computational model chosen for this study builds upon the model for detecting biological motion described in \cite{vignolo2016complexity}. The model takes inspiration from the human ability to distinguish between biological and non-biological motion, an ability exhibited by newborns where they orient their attention towards biologically moving stimuli \cite{simion2008predisposition}. The model exploits human motor movement regularities resulting from the \textit{Two-thirds power law}, a well known invariant of human movement \cite{viviani1992biological,richardson2002comparing} and has also been implemented on an iCub humanoid robot as a proof of applicability \cite{vignolo2016complexity,VignoloHumanoids2016,VignoloFrontiers2017}.   

The model for recognizing action similarities utilizes visual motion primitives to understand actions \cite{VignoloICPR2020}. The approach is to identify necessary and sufficient action sub-components and use them as visual primitives to form simple motion representations that can reconstruct a wide range of complex actions. A broad break down of the model's build is the following:

Firstly, the optical flow from the videos (of hand action stimuli) is extracted for each time instant, and the tangential velocity is computed (see \cite{vignolo2016complexity}). The averaged velocities over time give a compact representation of each video. The velocity sequences over time are segmented into sub-movements (portions). The sub-movements are derived automatically with set points that correspond to a Start, Stop, Change in the action dynamics, that are the local minima of the velocity profile \cite{Rea2019}. 

Secondly, the obtained sub-movements of all the actions (19 hand actions) are treated together and given as input to a K-means clustering, thereby building a unique dictionary of \textit{K} atoms. With the dictionary, each sub-movement of the training set is then reconstructed as an approximation of a linear combination of some of the atoms in the dictionary, using the sparse coding technique and represented as the sequence of weights used for each atom in the reconstruction. At the end of this procedure, given a video representing a given action, the model can describe each sub-movement \textit{u\textsubscript{i}} as the feature vector [\textit{$u_{i}^{1}$}, \textit{$u_{i}^{2}$}, ...\textit{$u_{i}^{K}$}], where \textit{$u_{i}^{j}$} is the coefficient/weight assigned to each atom (\textit{j}-th atom, where \textit{j} = 1...\textit{K}). Since the representation is sparse, some of the coefficients are equal to 0, and K=15 is the number of atoms of the dictionary. 

\begin{figure}[ht]
\begin{center}
\includegraphics[width=\linewidth]{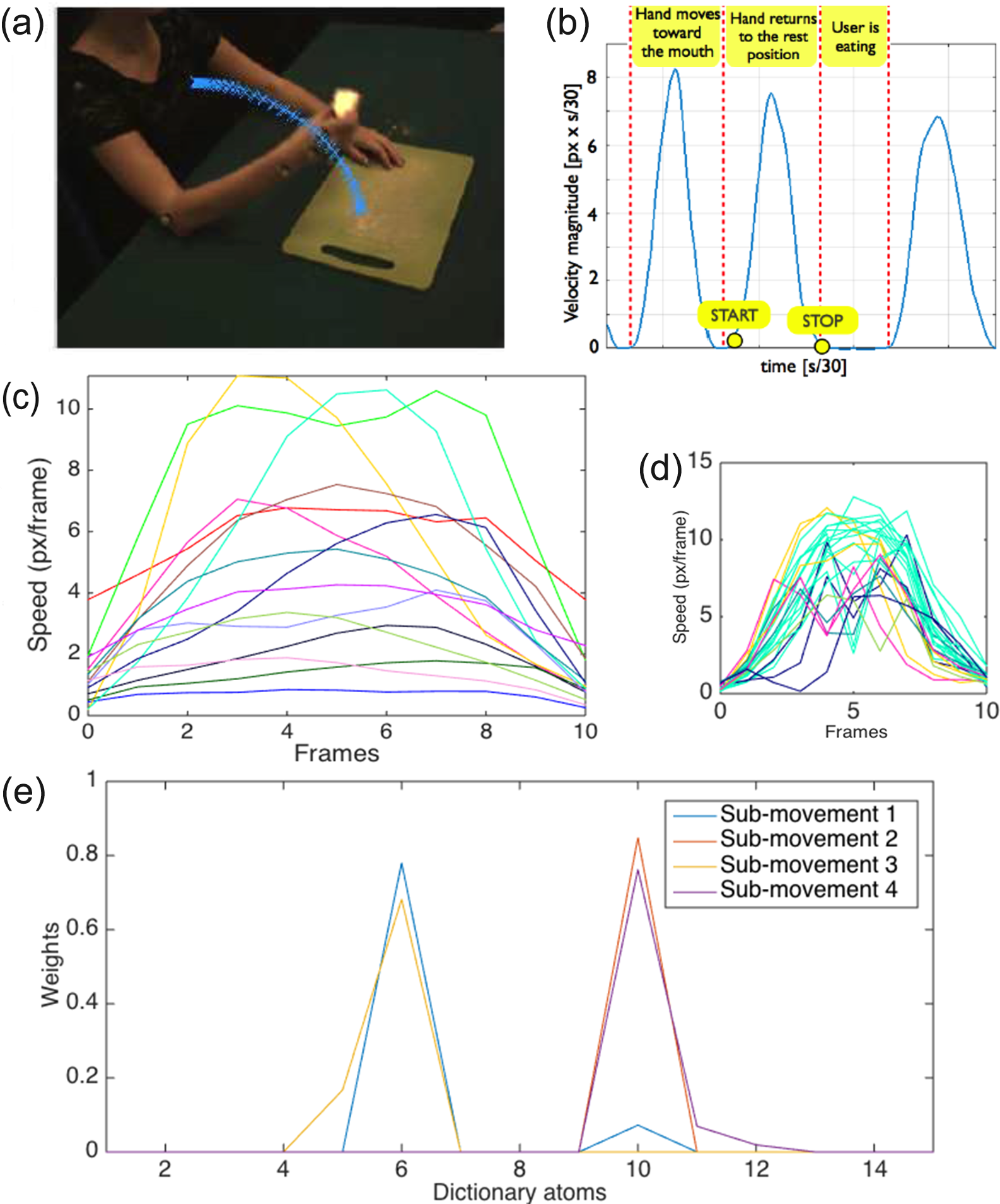}
\caption{ (a) \textit{Eat} action video from which optical flow is extracted, (b) Identified dynamic instants of \textit{Eat} action based on set rules and extracted sub-movements, (c) Dictionary of primitives composed of 15 sub-movements (atoms) extracted from all the 19 actions, (d) Sub-movements extracted from the \textit{Eat} action and (e) \textit{Transport} action represented via the dictionary primitives- the sub-movement 1 has a large contribution from the atom 6, sub-movement 2 has a large contribution from atom 10 and so on. Images modified from \cite{Rea2019} and \cite{VignoloICPR2020}.}

\label{fig:3d}
\end{center}
\end{figure}

Thirdly, a classification of the actions (19 hand actions) is performed following a supervised approach. A multi-class classifier is built with a one-vs-all approach, where a binary classifier per class (i.e., per action) is built. So for each action, a binary classifier is trained to discriminate between the representation of that action versus all the rest. See Fig. \ref{fig:3d} to see an example of how \textit{Eat} contributed to the sub-movement dictionary and how \textit{Transport} is represented via the dictionary primitives.

\section{Experiment 1}
This experiment addresses the extent to which the computational model based on kinematic primitives can determine action similarity among a group of hand-arm actions. It also addresses the extent to which the performance of the model relates to human similarity judgments of the same actions. A two-alternative forced-choice AST is designed for both the model and the human participants. The model and human participants perform the same task; hence the treatment procedures are designed to compare results. The comparison is made based on the accuracy, false-hit (incorrectly picked as target), and selection-bias (bias of an action classifier/choice actions to get picked).

\subsection{Model- Action similarity task}

\subsubsection{Stimuli}
For a given trial, video of the target action was fed to the model, and the model extracted optical flow from the video and computed the motion descriptor for each frame as described in \ref{computationalmodel}. The frontal viewpoint was used for both training and testing the model.

\subsubsection{Procedure}
There are 19 action classifiers trained on each of the chosen 19 actions. For the classification, we used a {\it Regularized Least Squares} (RLS) classifier, adopting the library GURLS \cite{tacchetti2012gurls} for an efficient implementation of RLS. We employ {\it Radial Basis Function} (RBF) as a kernel. The model performed an AST where it was presented with the target (T) action video and two action classifiers (A and B). These two classifiers competed to see which one of them (A or B) was the same as T. So for a given trial, where \textit{Eggs} is the T, then two action classifiers trained on say \textit{Eggs} (H) and \textit{Rolling} (M) shall compete, and the classifier with the higher score wins the trial. To simulate the constraints of a viewing period that a human participant would have, random instances of the stimuli were considered, where an instance is one sub-movement of the action (e.g., in the case of \textit{Mixing}, one half-circular rotation of the palm would be considered one sub-movement). The similarities were computed by averaging the similarities between 10 random instances of the actions.

\begin{figure}[ht]
\begin{center}
\includegraphics[width=\linewidth]{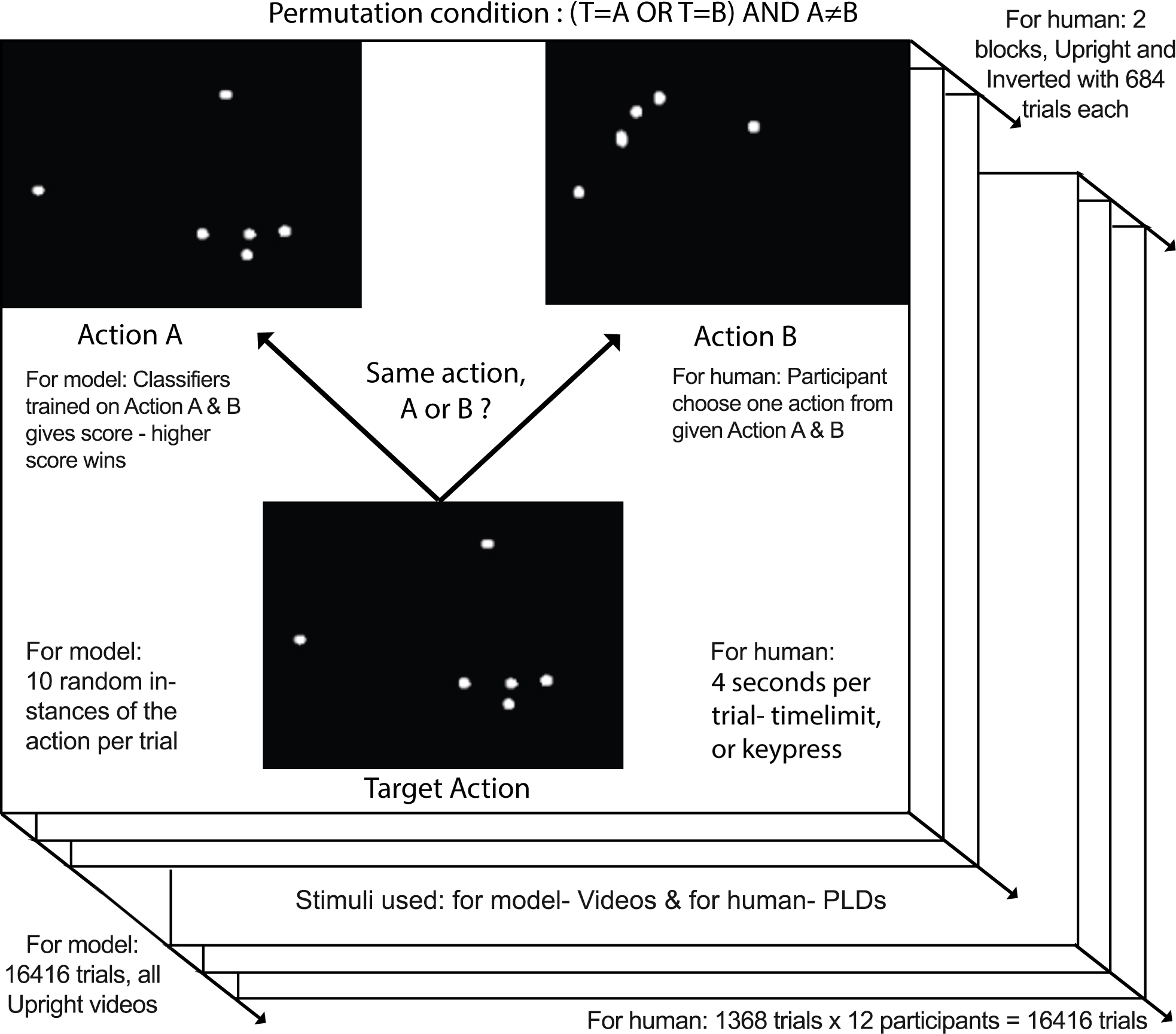}
\caption{ Schematic diagram of the experiment design for both the model and human AST. }
\label{fig:exp_design}
\end{center}
\end{figure}

See Fig. \ref{fig:exp_design} for a schematic description of the experiment design. Each trial consisted of the triad A, B, and T, with the condition (T=A OR T=B) AND A$\neq$B, i.e., one of the classifiers (A or B) always belongs (i.e., trained) to the same action as T. Therefore unique permutations = 684 (3(r) actions at a time taken from a set of 19(n) actions, with the order and repetition factor). The total trials conducted were 684x24 = 16416 in randomized order.

\subsection{Human- Action similarity task}
The human participants performed an equivalent version of the AST. PLDs of the actions were used, with no contextual information of the action (the tool used or the setting) provided -- limiting perceptual conditions to kinematic features. Additionally, to assess the participants' implicit semantic access, we tested their performance as a function of orientation: upright (UP) and inverted (INV) PLDs. If participants perform significantly poorly for the INV PLDs in contrast to the UP PLDs (inversion effect), that would indicate implicit semantic access for the UP PLDs.

\subsubsection{Participants}

Twelve subjects (5 males, mean age of 31.4 years, age range 24 to 46 years) with normal (or corrected) vision participated. They were provided information about the task, and gave written informed consent for participation. They were given a movie ticket for their participation time. The experiment was carried out in accordance with the National Ethics Law and the World Medical Association Declaration of Helsinki.

\subsubsection{Stimuli}
PLDs of the right arm for each of the actions (motion capture data) were generated using Biomotion toolbox-2 \cite{van2013biological} in MATLAB. The PLDs consist of six dots positioned at the shoulder, elbow, wrist, and three at the palm region (Fig. \ref{fig:stimuli}). Two orientations of PLDs were used: UP and INV (by horizontal flipping of UP PLD). See Fig. \ref{fig:exp_design} with a trial display of 3 PLDs (namely A, B, and T). The stimuli were presented in a frontal point of view (facing the participants) and played at their veridical speed. The experiment was conducted using MATLAB R2014a with Biomotion toolbox-2 \cite{van2013biological} and Psychtoolbox-3 \cite{kleiner2007s}. The stimuli were displayed on a 22 inch HP L2245wg LCD monitor, with a native resolution of 1680 x 1050 at 60 Hz, viewable dimension 29.5cm x 47.5cm(W x H), and a viewing distance of 100cm.

\subsubsection{Procedure}

Participants performed the same AST as the model in which they viewed three actions (A, B, and T) in one frame, and they had to indicate (via keypress) which of the two stimuli A or B was the same as the T stimulus. Each trial lasted for 4 seconds only, and the participants had to respond within the same period. Upon failure to respond in 4s, the next trial started. Participants were informed about the PLD's corresponding physical features, the viewpoint, and the orientations, but no information about the actions themselves was provided -- just that they were performing day-to-day actions. The PLDs had random starting frames that played in a continuous loop at 30 FPS. Each response was followed by a fixation cross (0.23$^{\circ}$) at the center (500-700ms). After providing instructions, the participants performed practice trials (30 trials), followed by the experiment.  

The experiment consisted of 3 independent variables in a mixed design; Orientation (UP/INV, within-subjects), Block-order (UP-INV/INV-UP, between-subjects), and Actions (19 actions, random variable). See Fig. \ref{fig:exp_design} for a schematic description. The block-orders (UP-INV and INV-UP) were balanced between the subjects, with 6 participants viewing UP-INV. Individual trial orders within blocks were randomized. The overall trials performed were the same as the model (16416 trials).

\subsection{Results}
\begin{figure*}[!h]
\begin{center}
\includegraphics[width=\textwidth]{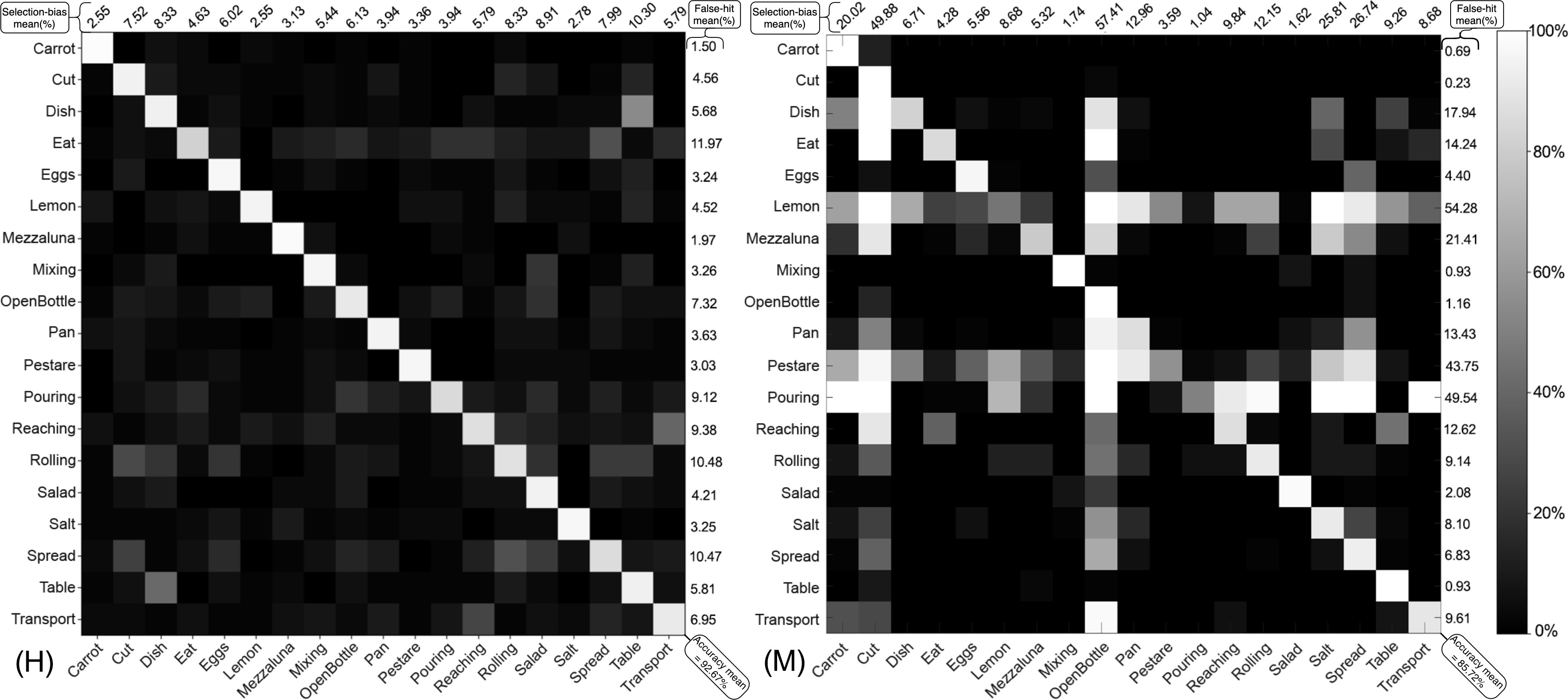}
\caption{ Matrices with mean similarity measures, with target actions (y-axis) and matched actions or classifiers (x-axis). Measures for (H) Human (matches in \%) and (M) Model (matches in \%).}
\label{fig:matrices-figure}
\end{center}
\end{figure*}

The model's and the human participants' performance are presented in the form of confusion matrices for humans (H) and the model (M) in Fig. \ref{fig:matrices-figure}. Each matrix shows the similarity measure (accuracy of matching rate\%) along the matrix's diagonal. False-hit (frequency) for the target actions is shown at the end of their respective rows. Selection-bias (frequency) for the action/classifiers is shown at the end of their respective columns. These measures in the matrices highlight the strengths and weaknesses of the performer (model or human) for each of the actions (or action classifiers).

The confusion matrices (H) and (M) show the matching rate\%, with target actions on the y-axis and matched actions (classifiers for (M)) on the x-axis. A cell$_{(i, j)}$ has the match\% of the \textit{i$^{th}$} target action matched with the \textit{j$^{th}$} action. The diagonal cells$_{(i=j)}$ (referred to as the accuracy cells) indicate the percentage of times in which the target action was correctly identified. For example, matrix H, cell$_{(3,3)}$ = 94.1\%, shows the times the target action \textit{Dish} was identified correctly.

The non-diagonal cells$_{(i \neq j)}$ report the \% of times the target action was incorrectly identified, which is split into false-hits and selection-bias. The similarity measures (\%) in the $row_i$ are averaged together minus the respective accuracy cell, indicating the false-hit\% of the \textit{i$^{th}$} target action, e.g., for $row_6$ in matrix M, cell$_{(6,1)}$ = 62.5\% of the times \textit{Lemon} was identified as \textit{Carrot}, whenever \textit{Carrot} was the other classifier/action and the $row_6$ averages 54.28\% (excluding cell$_{(6,6)}$), indicating \textit{Lemon}'s mean \textit{false-hits\%}. 
The similarity measures (\%) in the $column_j$ are averaged together minus the accuracy cell, indicating the \textit{selection-bias\%} of the \textit{j$^{th}$} action/classifier. E.g., for $column_2$ in matrix M, cell$_{(1,2)}$ = 12.5\% of the times \textit{Cut} action/classifier selected as target action \textit{Carrot} and the $column_2$ averages 49.88\% (excluding cell$_{(2,2)}$), indicating \textit{Cut}'s mean \textit{selection-bias\%}.

Independent sample t-tests were conducted on accuracy, false-hit and selection-bias measures for the two matrices. Both the model (\textit{M} = 85.72\%, \textit{SD} = 16.80) and the results for the human participants (\textit{M} = 92.67\%, \textit{SD} = 4.82) showed a high level of accuracy with no significant difference between them; \textit{t}(36) = -1.73, \textit{p} = 0.092. Although the false-hit results were relatively low for both matrices, the model had significantly higher false-hit rates (\textit{M} = 14.28\%, \textit{SD} = 16.80) than the human participants (\textit{M} = 5.81\%, \textit{SD} = 3.16); \textit{t}(36) = -2.16, \textit{p} = 0.037. Similarly the selection-bias results were also relatively low for both matrices, but the model had a significantly higher selection-bias (\textit{M} = 14.28\%, \textit{SD} = 15.78) than the human participants (\textit{M} = 5.65\%, \textit{SD} = 2.38); \textit{t}(36) = -2.16, \textit{p} = 0.023.

 \begin{table}[!ht]
\begin{center} 
\caption{RT means(in seconds),and (SD) across conditions} 
\label{Means} 
\includegraphics[width=\linewidth]{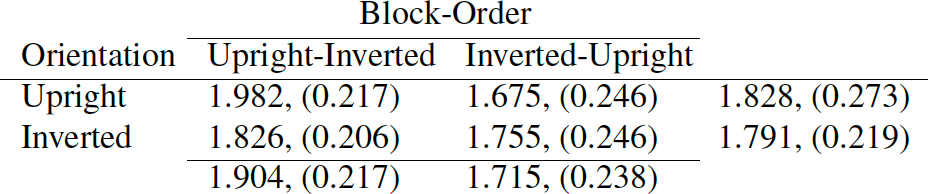}
\end{center} 
\end{table}

\vspace{0pt}
Concerning the second question on how the model's performance relates to human action similarity judgments, a crucial part of which is to show the extent to which humans relied on the kinematic features to make their similarity judgments. If human action similarity judgments are not affected by the actions' orientation, this would indicate that kinematic features were used as the basis for the judgments -- lack of semantic level access for INV displays that would normally lead to a difference in performance between UP and INV displays \cite{hemeren2011deriving}.

A 2 \textit{orientation} (within-subject) x 2 \textit{block-order} (between-subject) mixed ANOVA was performed on the accuracy and reaction time (RT, correct responses only) measures for the human-AST data (all subjects were included). The actions were treated as random variables. The RT results are presented in Table~\ref{Means}. The main effect of orientation was not significant (\textit{F}(1,11) = 1.197, $\eta_{p}^{2}$ = 0.107, \textit{p} = 0.299). There is no performance difference between UP and INV action stimuli. The main effect of block-order was also not significant (\textit{F}(1,11) = 2.175, $\eta_{p}^{2}$ = 0.179, \textit{p} =0.171). There was however a significant interaction effect (\textit{F}(1,11) = 11.585, $\eta_{p}^{2}$ = 0.537, \textit{p} = 0.007). The significant difference leading to the interaction effect consists of faster responses for UP displays (\textit{M} = 1.675s, \textit{SD} = 0.246) when presented after INV displays(\textit{M} = 1.982s, \textit{SD} = 0.217); \textit{t}(10) = 2.31, \textit{p} = 0.043. Further analyses between UP and INV did not show any significant differences, no simple main effects for orientation (\textit{p} $>$ 0.05). Regarding accuracy, human participants performed equally well for both \textit{UP} (\textit{M} = 92.9\%, \textit{SD} = 3.29) and \textit{INV} (\textit{M} = 92.4\%, \textit{SD} = 3.65) conditions with no significant main effects or interaction effect. 

\subsection{Discussion}

From the accuracy measure, both the model and human participants performed reliably well with no significant difference. However, the model has certain drawbacks compared to participants in terms of overall performance, i.e., the model has significantly more false-hits and a significantly greater selection bias. That said, the observed differences come from a small set of action classifiers and target actions. Here we examine those cases to see for a possible cause.

For the selection-biases, action classifiers (matrix M) for \textit{Openbottle} (57.41\%), \textit{Cut} (49.88\%), and \textit{Spread} (26.74\%) (in order of decreasing measure), in cases when they are pitted against another target action -- get selected instead of the correct one. These actions have the most number of kinematic primitives (atoms) that make up the dictionary primitives, so in a way, these actions contain most of the primitives that make up the sub-movements of all the 19 actions. Thereby these actions correspond also to other actions with populated sub-movements forming different atoms. Hence they have more chances to get confused with other actions, thereby leading to a high selection-bias. This also explains why the action classifiers with a high selection-bias have higher accuracy also, as they have sufficient primitives to create a strong representation of their own action.

Concerning the false-hits, target actions (matrix M) such as \textit{Lemon} (54.28\%), \textit{Pouring} (49.54\%), and \textit{Pestare} (43.75\%) (in order of decreasing measure), got the most number of false-hits along with low accuracy. These actions show a lack of descriptive capability, i.e., poor representation of the action by the dictionary primitives. This is in addition to their respective classifiers getting a low selection-bias, also pointing towards a lack of sufficient kinematic primitives. False-hits for these actions may result from the classifiers' training process -- that necessary and sufficient primitives were not extracted properly for the dictionary. Nevertheless, further studies will be needed to confirm these considerations, specifically whether a) the model performance can be improved by increasing the number of dictionary atoms (K), b) the training can be improved with better action videos, better as in longer temporal sequence, or different viewpoints. 

Regarding the low selection-bias and low false-hits for humans, a possibility is that they were relying on action semantics to aid their judgment. In AST, we probed for implicit access to action semantics through orientation manipulation, with no difference. These results rule out semantic level access for UP displays. To further affirm that the participants had no idea what the actions were (at least to the point to aid them in AST), we conducted Experiment 2 to test for explicit access to action semantics. 

\section{Experiment 2}

This experiment addresses the third question on whether the human judgments in AST were based solely on the kinematic features of the actions. A five-alternative forced-choice AIT is presented to human participants (no participants from Experiment 1), where they had to identify the displayed action from a list of five action labels.

\subsection{Human- Action identification task}

\subsubsection{Participants}
Fifty-four Mechanical Turk workers (33 males, mean age of 37.33 years, age range 26 to 73 years) with normal (or corrected) vision and fluent in English participated. They were informed about the task and provided informed consent for their participation. Participants received monetary compensation of \$2.50 for their participation time. The experiment was carried out in accordance with the National Ethics Law and the World Medical Association Declaration of Helsinki.

\subsubsection{Stimuli}
The trial display consisted of one action PLD at a time followed by five action labels. The PLDs (19 actions) were the same as in Experiment 1: frontal viewpoint played at veridical speed with UP and INV orientation. The stimuli were displayed using Amazon Mechanical Turk with extensions from psiTurk \cite{gureckis2016psiturk} and jsPsych \cite{de2015jspsych}. 

\subsubsection{Procedure}
Participants performed an AIT where they were shown an action (target) for 4 seconds, after which they had to identify (mouse click) the target action label from 5 action labels (alternatives) within 10 seconds. The alternatives consisted of the correct label, and four randomly chosen (from the same pool of 19 action labels) labels with no repetition. Clicking or failing to respond within 10 seconds led to the next trial (preceded by a fixation cross for 700ms). The display orientation (UP or INV) was informed prior to the start. Participants were informed of the PLDs (identical to Experiment 1-human AST). The instructions were on-screen with example displays. After the instructions, a video of a trial was shown (no practice session). There were questionnaires about the difficulty of the task at the end of the experiment.

The experimental design is identical to Experiment 1-human AST. The block-order (UP-INV and INV-UP) were balanced between the subjects, with 29 participants viewing INV-UP. Individual trial orders were randomized for each participant. The blocks had 19 trails where each trial presented one of the 19 actions; the total number of trials per participant was 38. 

\subsection{Results}
\begin{figure}[ht]
\begin{center}
\includegraphics[width=\linewidth]{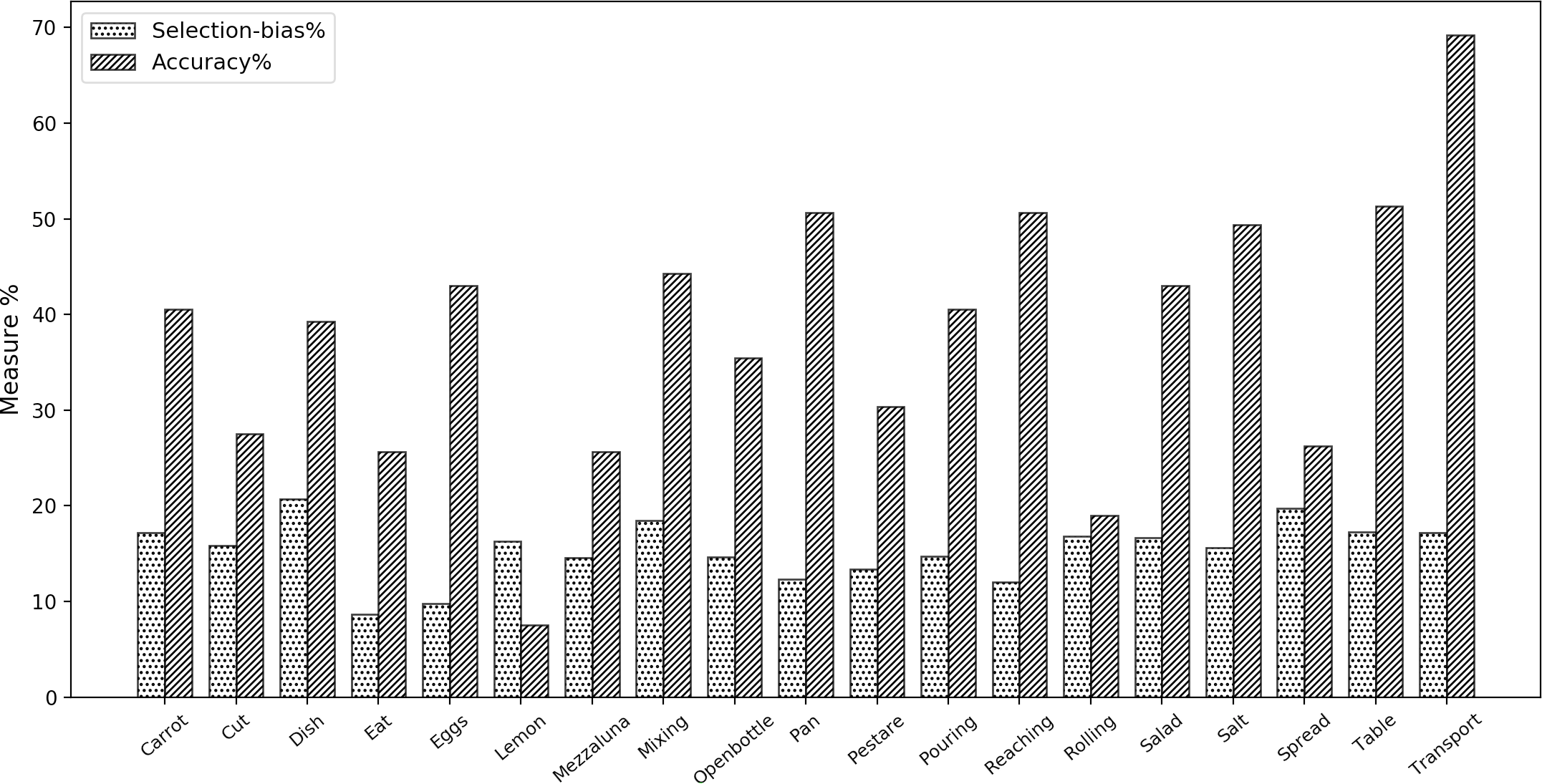}
\caption{ Accuracy(\%) and selection-bias\% for Experiment 2}
\label{fig:bar}
\end{center}
\end{figure}

We had a selection criterion where the participant's mean RT should exceed 2 seconds; this was to ensure that the participants diligently performed the task. Therefore 14 participants were excluded, and data from 40 were taken for the analysis. Fig. \ref{fig:bar} shows the accuracy\% (for correct identification) and the selection bias\%. To confirm humans' reliance on kinematic features for their similarity judgments -- we had to rule out explicit semantic level access for the PLDs. If participants perform poorly in identifying the PLDs, irrespective of the display orientation, this would strongly suggest limited semantic level access.

The overall accuracy (\textit{M} = 37.85\%, \textit{SD} = 14.17) indicates poor performance with a mean selection-bias of 15.35\% (\textit{SD} = 3.12).  Participants performed poorly for both UP displays (\textit{M} = 38.68\%, \textit{SD} = 15.61) and INV displays (\textit{M} = 35.92\%, \textit{SD} = 16.25). A 2 \textit{orientation} (within-subject) x 2 \textit{block-order} (between-subject) mixed ANOVA was performed on the accuracy to check for an inversion effect. The actions were treated as random variables. There was no significant main effect of orientation ( \textit{F}(1,39) = 0.966, $\eta_{p}^{2}$ = 0.025, \textit{p} = 0.332), indicating no performance difference between UP and INV action stimuli. The main effect of block-order was also not significant (\textit{F}(1,39) = 1.807, $\eta_{p}^{2}$ = 0.045, \textit{p} = 0.187). 
 There was however a significant interaction effect (\textit{F}(1,39) = 6.152, $\eta_{p}^{2}$ = 0.139, \textit{p} = 0.018). The significant difference leading to the interaction effect consists of higher accuracy for responses for INV displays (\textit{M} = 29.74\%, \textit{SD} = 14.01)  when presented after UP displays (\textit{M} = 42.11, \textit{SD} = 16.29); \textit{t}(38) = 2.57, \textit{p} = 0.014. 
 

\subsection{Discussion}

Experiment 2 shows a poor overall accuracy(\%), indicating that the participants were having difficulty identifying the actions from the displayed PLDs. Although most of the actions were identified above chance level (i.e., 20\%, out of 5 options), very few actions had a relatively high accuracy such as \textit{Transport = 69\%, Reaching = 50\%} and \textit{Table= 50\%}. Despite the poor accuracy, there was no particular selection-bias pattern. The kinematic information within the PLDs may not be enough for the participants to recognize the action and choose the correct action labels, which also points to why they did not show any particular selection preference.

Observing the results of AIT in light of AST, no inversion effect was observed for both the tasks and the poor accuracy in AIT indicate that the participants were not relying on semantics in the AST. Hence we show that humans did rely mainly on the kinematic features of the actions to perform AST -- similar to the model.

\section{General Discussion}

In this work, a comparison between a computational model's performance and human judgments was carried out by using a common task -- to understand the visual processing of action similarities better. To this purpose, we designed a similarity judgment task using the Multiview Cooking Actions dataset and considered different research questions on the reliability of the computational model and its similarity with human observers' choices. 

Overall, both the model and human participants could reliably identify whether given actions were the same or not, which indicates that the model and humans might be using similar information -- an aspect which is the objective of our current, deeper investigation. 

In our first experiment, human performance was better than the model in terms of low selection-bias and low false-hits. Given the very simple description adopted in the model, which is entirely based on low-level kinematic features, with no integration of information over time, this dissimilarity in performance is relatively limited. To ensure that this difference was not mainly due to the fact that humans exploit action semantics to aid their judgment, we performed the same experiment with INV stimuli, and we conducted an action identification study. The results of both analyses indicated that it is unlikely that semantics had been used. 

An aspect deserving further attention refers to the differences in the type of information provided to the computational model and humans, which may be partially the cause of the differences in the performance. To address this issue, we are currently performing an investigation in which the same computational method is applied to motion capture data. Additionally, further investigation is needed to understand whether humans utilize kinematic primitives to judge action similarities -- if so, how and is it in the same manner as the model.

The current work provides an insight into the potential mechanisms supporting action similarity detection in humans, providing a pathway towards implementing similar models in machines. 
The approach has a developmental inspiration, in that it builds upon an existing model of newborns' ability (biological motion detection \cite{VignoloFrontiers2017}) to assess how far such a simple representation allows to go in terms of a novel, more complex skill as the detection of action similarity.
It is important to note that progressive development could continue from there toward more complex social competences. 
In fact, for human beings detecting action similarity plays a fundamental role in imitation. In particular, according to the similarity model \cite{hale2016cognitive} kinematic similarity increases the predictability of the action.
Imitation, in turn, supports the development of action understanding.
For instance, several researchers have suggested that the experience of being imitated is crucial in the development of the Mirror Neurons System (e.g., \cite{Catmur2009,Jones2006InfantsLT}).
In this context, the child's ability to judge the kinematic similarity between her and her caregiver's actions would support the child's ability to mimic, a further step towards action understanding. 

In a similar vein, the topic of imitation has been widely investigated also in robotics (e.g., \cite{Kuniyoshi2003FromVS,NagaiMNS2011,BillHellLej2011,billing2010behavior}) and bears important implications for the domain of learning from demonstration \cite{Argall2009}.
Additionally, for this application, the possibility of detecting action similarity and performing actions that closely resembles that of the human partner could increase the intuitiveness and efficacy of the interaction.\\

\section*{Acknowledgment}
This work has been partially carried out at the Machine Learning Genoa (MaLGa) center, Università di Genova (IT).\\ \\
\small
\section*{Publication Note}
 This paper is a pre-publication draft of the contribution to appear as part of Proceedings of the 10th Joint IEEE International Conference on Development and Learning and Epigenetic Robotics (ICDL-EpiRob 2020),  26-30th October, 2020, Valpara\'iso, Chile. \url{https://cdstc.gitlab.io/icdl-2020/}.\\

\balance
\bibliographystyle{IEEEtranN}
\bibliography{biblography}

\begin{thebibliography}{34}
\providecommand{\natexlab}[1]{#1}
\providecommand{\url}[1]{#1}
\csname url@samestyle\endcsname
\providecommand{\newblock}{\relax}
\providecommand{\bibinfo}[2]{#2}
\providecommand{\BIBentrySTDinterwordspacing}{\spaceskip=0pt\relax}
\providecommand{\BIBentryALTinterwordstretchfactor}{4}
\providecommand{\BIBentryALTinterwordspacing}{\spaceskip=\fontdimen2\font plus
\BIBentryALTinterwordstretchfactor\fontdimen3\font minus
  \fontdimen4\font\relax}
\providecommand{\BIBforeignlanguage}[2]{{%
\expandafter\ifx\csname l@#1\endcsname\relax
\typeout{** WARNING: IEEEtranN.bst: No hyphenation pattern has been}%
\typeout{** loaded for the language `#1'. Using the pattern for}%
\typeout{** the default language instead.}%
\else
\language=\csname l@#1\endcsname
\fi
#2}}
\providecommand{\BIBdecl}{\relax}
\BIBdecl

\bibitem[Tversky(2019)]{tversky2019mind}
B.~Tversky, \emph{Mind in motion: How action shapes thought}.\hskip 1em plus
  0.5em minus 0.4em\relax Hachette UK, 2019.

\bibitem[Yovel and O'Toole(2016)]{yovel2016recognizing}
G.~Yovel and A.~J. O'Toole, ``Recognizing people in motion,'' \emph{Trends in
  cognitive sciences}, vol.~20, no.~5, pp. 383--395, 2016.

\bibitem[Simion et~al.(2008)Simion, Regolin, and
  Bulf]{simion2008predisposition}
F.~Simion, L.~Regolin, and H.~Bulf, ``A predisposition for biological motion in
  the newborn baby,'' \emph{Proceedings of the National Academy of Sciences},
  vol. 105, no.~2, pp. 809--813, 2008.

\bibitem[Sifre et~al.(2018)Sifre, Olson, Gillespie, Klin, Jones, and
  Shultz]{sifre2018longitudinal}
R.~Sifre, L.~Olson, S.~Gillespie, A.~Klin, W.~Jones, and S.~Shultz, ``A
  longitudinal investigation of preferential attention to biological motion in
  2-to 24-month-old infants,'' \emph{Scientific reports}, vol.~8, no.~1, pp.
  1--10, 2018.

\bibitem[Hemeren(2008)]{hemeren2008mind}
P.~E. Hemeren, ``Mind in action,'' \emph{Lund University Cognitive Studies},
  vol. 140, 2008.

\bibitem[Watson and Buxbaum(2014)]{watson2014uncovering}
C.~E. Watson and L.~J. Buxbaum, ``Uncovering the architecture of action
  semantics.'' \emph{Journal of Experimental Psychology: Human Perception and
  Performance}, vol.~40, no.~5, p. 1832, 2014.

\bibitem[Giese and Lappe(2002)]{giese2002measurement}
M.~Giese and M.~Lappe, ``Measurement of generalization fields for the
  recognition of biological motion,'' \emph{Vision research}, vol.~42, no.~15,
  pp. 1847--1858, 2002.

\bibitem[Catmur and Heyes(2013)]{catmur2013you}
C.~Catmur and C.~Heyes, ``Is it what you do, or when you do it? the roles of
  contingency and similarity in pro-social effects of imitation,''
  \emph{Cognitive Science}, vol.~37, no.~8, pp. 1541--1552, 2013.

\bibitem[Kong and Fu(2018)]{kong2018human}
Y.~Kong and Y.~Fu, ``Human action recognition and prediction: A survey,''
  \emph{arXiv preprint arXiv:1806.11230}, 2018.

\bibitem[Qin et~al.(2015)Qin, Liu, Zhang, Wang, and Shao]{qin2015compressive}
J.~Qin, L.~Liu, Z.~Zhang, Y.~Wang, and L.~Shao, ``Compressive sequential
  learning for action similarity labeling,'' \emph{IEEE Transactions on Image
  Processing}, vol.~25, no.~2, pp. 756--769, 2015.

\bibitem[Kliper-Gross et~al.(2011)Kliper-Gross, Hassner, and
  Wolf]{kliper2011action}
O.~Kliper-Gross, T.~Hassner, and L.~Wolf, ``The action similarity labeling
  challenge,'' \emph{IEEE Transactions on Pattern Analysis and Machine
  Intelligence}, vol.~34, no.~3, pp. 615--621, 2011.

\bibitem[Vignolo et~al.(2016{\natexlab{a}})Vignolo, Noceti, Sciutti, Rea,
  Odone, and Sandini]{vignolo2016complexity}
A.~Vignolo, N.~Noceti, A.~Sciutti, F.~Rea, F.~Odone, and G.~Sandini, ``The
  complexity of biological motion,'' in \emph{2016 Joint IEEE International
  Conference on Development and Learning and Epigenetic Robotics
  (ICDL-EpiRob)}.\hskip 1em plus 0.5em minus 0.4em\relax IEEE, 2016, pp.
  66--71.

\bibitem[Hemeren and Thill(2011)]{hemeren2011deriving}
P.~E. Hemeren and S.~Thill, ``Deriving motor primitives through action
  segmentation,'' \emph{Frontiers in psychology}, vol.~1, p. 243, 2011.

\bibitem[Kuli{\'c} et~al.(2011)Kuli{\'c}, Kragic, and
  Kr{\"u}ger]{kulic2011learning}
D.~Kuli{\'c}, D.~Kragic, and V.~Kr{\"u}ger, ``Learning action primitives,'' in
  \emph{Visual analysis of humans}.\hskip 1em plus 0.5em minus 0.4em\relax
  Springer, 2011, pp. 333--353.

\bibitem[Malafronte et~al.(2017)Malafronte, Goyal, Vignolo, Odone, and
  Noceti]{malafronte2017investigating}
D.~Malafronte, G.~Goyal, A.~Vignolo, F.~Odone, and N.~Noceti, ``Investigating
  the use of space-time primitives to understand human movements,'' in
  \emph{International Conference on Image Analysis and Processing}.\hskip 1em
  plus 0.5em minus 0.4em\relax Springer, 2017, pp. 40--50.

\bibitem[Vignolo et~al.(2020)Vignolo, Noceti, Sciutti, Odone, and
  Sandini]{VignoloICPR2020}
A.~Vignolo, N.~Noceti, A.~Sciutti, F.~Odone, and G.~Sandini, ``Learning
  dictionaries of kinematic primitives for action classification,''
  \emph{International Conference on Pattern Recognition (ICPR)}, 2020.

\bibitem[Viviani and Stucchi(1992)]{viviani1992biological}
P.~Viviani and N.~Stucchi, ``Biological movements look uniform: evidence of
  motor-perceptual interactions.'' \emph{Journal of experimental psychology:
  Human perception and performance}, vol.~18, no.~3, p. 603, 1992.

\bibitem[Richardson and Flash(2002)]{richardson2002comparing}
M.~J. Richardson and T.~Flash, ``Comparing smooth arm movements with the
  two-thirds power law and the related segmented-control hypothesis,''
  \emph{Journal of neuroscience}, vol.~22, no.~18, pp. 8201--8211, 2002.

\bibitem[Vignolo et~al.(2016{\natexlab{b}})Vignolo, Rea, Noceti, Sciutti,
  Odone, and Sandini]{VignoloHumanoids2016}
A.~Vignolo, F.~Rea, N.~Noceti, A.~Sciutti, F.~Odone, and G.~Sandini,
  ``Biological movement detector enhances the attentive skills of humanoid
  robot icub,'' in \emph{IEEE-RAS International Conference on Humanoid Robots},
  2016.

\bibitem[Vignolo et~al.(2017)Vignolo, Noceti, Rea, Sciutti, Odone, and
  Sandini]{VignoloFrontiers2017}
A.~Vignolo, N.~Noceti, F.~Rea, A.~Sciutti, F.~Odone, and G.~Sandini,
  ``Detecting biological motion for human-robot interaction: A link between
  perception and action,'' \emph{Frontiers in Robotics and AI}, 2017.

\bibitem[Rea et~al.(2019)Rea, Vignolo, Sciutti, and Noceti]{Rea2019}
F.~Rea, A.~Vignolo, A.~Sciutti, and N.~Noceti, ``Human motion understanding for
  selecting action timing in collaborative human-robot interaction,''
  \emph{Frontiers in Robotics and AI}, 2019.

\bibitem[Tacchetti et~al.(2012)Tacchetti, Mallapragada, Santoro, and
  Rosasco]{tacchetti2012gurls}
A.~Tacchetti, P.~S. Mallapragada, M.~Santoro, and L.~Rosasco, ``Gurls: a
  toolbox for regularized least squares learning,'' 2012.

\bibitem[van Boxtel and Lu(2013)]{van2013biological}
J.~J. van Boxtel and H.~Lu, ``A biological motion toolbox for reading,
  displaying, and manipulating motion capture data in research settings,''
  \emph{Journal of vision}, vol.~13, no.~12, pp. 7--7, 2013.

\bibitem[Kleiner et~al.(2007)Kleiner, Brainard, and Pelli]{kleiner2007s}
M.~Kleiner, D.~Brainard, and D.~Pelli, ``What's new in psychtoolbox-3?'' 2007.

\bibitem[Gureckis et~al.(2016)Gureckis, Martin, McDonnell, Rich, Markant,
  Coenen, Halpern, Hamrick, and Chan]{gureckis2016psiturk}
T.~M. Gureckis, J.~Martin, J.~McDonnell, A.~S. Rich, D.~Markant, A.~Coenen,
  D.~Halpern, J.~B. Hamrick, and P.~Chan, ``psiturk: An open-source framework
  for conducting replicable behavioral experiments online,'' \emph{Behavior
  research methods}, vol.~48, no.~3, pp. 829--842, 2016.

\bibitem[De~Leeuw(2015)]{de2015jspsych}
J.~R. De~Leeuw, ``jspsych: A javascript library for creating behavioral
  experiments in a web browser,'' \emph{Behavior research methods}, vol.~47,
  no.~1, pp. 1--12, 2015.

\bibitem[Hale and Hamilton(2016)]{hale2016cognitive}
J.~Hale and A.~F. d.~C. Hamilton, ``Cognitive mechanisms for responding to
  mimicry from others,'' \emph{Neuroscience \& Biobehavioral Reviews}, vol.~63,
  pp. 106--123, 2016.

\bibitem[Catmur et~al.(2009)Catmur, Walsh, and Heyes]{Catmur2009}
C.~Catmur, V.~Walsh, and C.~Heyes, ``Associative sequence learning: the role of
  experience in the development of imitation and the mirror system,''
  \emph{PhilosophicalTransactions of the Royal Society B}, vol. 364, no. 1528,
  pp. 2369--2380, 2009.

\bibitem[Jones(2006)]{Jones2006InfantsLT}
S.~S. Jones, ``Infants learn to imitate by being imitated,'' in
  \emph{Proceedings of the International Conference on Development and
  Learning: The Tenth International Conference on Development and
  Learning}.\hskip 1em plus 0.5em minus 0.4em\relax Indiana University
  Bloomington, IN, 2006.

\bibitem[Yasuo et~al.(2003)Yasuo, Yasuaki, Masayuki, and
  Hirochika]{Kuniyoshi2003FromVS}
K.~Yasuo, Y.~Yasuaki, I.~Masayuki, and I.~Hirochika, ``From visuo-motor self
  learning to early imitation-a neural architecture for humanoid learning,''
  \emph{2003 IEEE International Conference on Robotics and Automation}, vol.~3,
  pp. 3132--3139 vol.3, 2003.

\bibitem[Nagai et~al.()Nagai, Kawai, and Asada]{NagaiMNS2011}
Y.~Nagai, Y.~Kawai, and M.~Asada, ``Emergence of mirror neuron system: Immature
  vision leads to self-other correspondence,'' in \emph{2011 IEEE International
  Conference on Development and Learning (ICDL)}, vol.~2, pp. 1--6.

\bibitem[Billing et~al.(2011)Billing, Hellstr{\"{o}}m, and
  Janlert]{BillHellLej2011}
E.~A. Billing, T.~Hellstr{\"{o}}m, and L.~E. Janlert, ``{Predictive learning
  from demonstration},'' in \emph{Agents and artificial Intelligence: Second
  International Conference, ICAART 2010, Valencia, Spain, January 22-24, 2010.
  Revised Selected Papers}.\hskip 1em plus 0.5em minus 0.4em\relax Springer
  Verlag, 2011, pp. 186--200.

\bibitem[Billing et~al.(2010)Billing, Hellstr{\"o}m, and
  Janlert]{billing2010behavior}
E.~A. Billing, T.~Hellstr{\"o}m, and L.-E. Janlert, ``Behavior recognition for
  learning from demonstration,'' in \emph{Proceedings of IEEE International
  Conference on Robotics and Automation}.\hskip 1em plus 0.5em minus
  0.4em\relax IEEE, 2010, pp. 866--872.

\bibitem[Argall et~al.(2009)Argall, Chernova, Veloso, and Browning]{Argall2009}
B.~Argall, S.~Chernova, M.~Veloso, and B.~Browning, ``A survey of robot
  learning from demonstration,'' \emph{Robotics and Autonomous Systems},
  vol.~57, pp. 469--483, 2009.

\end{thebibliography}

\end{document}